\documentclass[sigconf, nonacm]{acmart}
\AtBeginDocument{%
  }

\usepackage{algorithm}
\usepackage{algorithmic}
\usepackage{booktabs}
\usepackage{amsmath}
\usepackage{graphicx}
\usepackage{subfigure}
\usepackage{enumitem}

\begin{document}

\title{Retrieval-Augmented Multi-scale Framework for County-Level Crop Yield Prediction Across Large Regions}



\author{Yiming Sun}
\affiliation{%
  \institution{University of Pittsburgh}
  \city{Pittsburgh}
  \state{PA}
  \country{USA}
}
\email{yimingsun@pitt.edu}

\author{Qi Cheng}
\affiliation{%
  \institution{University of Pittsburgh}
  \city{Pittsburgh}
  \state{PA}
  \country{USA}
}
\email{qic69@pitt.edu}

\author{Licheng Liu}
\affiliation{%
  \institution{University of Wisconsin-Madison}
  \city{Madison}
  \state{WI}
  \country{USA}
}
\email{licheng.liu@wisc.edu}

\author{Runlong Yu}
\affiliation{%
  \institution{University of Alabama}
  \city{Tuscaloosa}
  \state{AL}
  \country{USA}
}
\email{ryu5@ua.edu}

\author{Yiqun Xie}
\affiliation{%
  \institution{University of Maryland}
  \city{College Park}
  \state{MD}
  \country{USA}
}
\email{xie@umd.edu}

\author{Xiaowei Jia}
\affiliation{%
  \institution{University of Pittsburgh}
  \city{Pittsburgh}
  \state{PA}
  \country{USA}
}
\email{xiaowei@pitt.edu}

\renewcommand{\shortauthors}{Sun et al.}

\begin{abstract}
  This paper proposes a new method for crop yield prediction, which is essential for developing management strategies, informing insurance assessments, and ensuring long-term food security. Although existing data-driven approaches have shown promise in this domain, their performance often degrades when applied across large geographic regions and long time periods. This limitation arises from two key challenges: (1) difficulty in jointly capturing short-term and long-term temporal patterns, and (2) inability to effectively accommodate spatial data variability in agricultural systems. Ignoring these issues often leads to unreliable predictions for specific regions or years, which ultimately affects policy decisions and resource allocation. In this paper, we propose a new predictive framework to address these challenges. First, we introduce a new backbone model architecture that captures both short-term daily-scale crop growth dynamics and long-term dependencies across years. To further improve generalization across diverse spatial regions, we augment this model with a retrieval-based adaptation strategy. 
  Recognizing the substantial yield variation across years, we design a novel retrieval-and-refinement pipeline that adjusts retrieved samples by removing cross-year bias not explained by input features. Our experiments on real-world county-level corn yield data over 630 counties in the US demonstrate that our method consistently outperforms different types of baselines. The results also verify the effectiveness of the retrieval-based augmentation method in improving model robustness under spatial heterogeneity.    
\end{abstract}

 \maketitle






\section{Introduction}

Agricultural monitoring is essential for understanding and managing the complex interactions between crop growth and the cycling of carbon, nitrogen, and water in agroecosystems.  Effective monitoring enables reliable estimates of crop yields, provides insights into soil health and environmental impacts, and facilitates the quantification of greenhouse gas emissions~\cite{smith2016global,becker2020strengthening,tian2020comprehensive}.  These capabilities are increasingly important in the context of global challenges such as climate change, resource scarcity, and the need for sustainable intensification of food production given the growing population.  

Although crop yield is often  reported for historical years (typically at the county level or state level), a key practical need is to provide accurate estimates of crop yield promptly by the end of each current year, once that year’s meteorological, soil, and management data are available. 
Such current-year prediction provides a principled way to generate  yield estimates in regions with missing or delayed reporting, support post-season assessment of how climate and management shaped production, and enable analysis of environmental impacts and cross-region comparisons of management scenarios. 
More importantly,  timely current-year estimates inform the development of policies and management planning, including next-season decisions (e.g., irrigation and nutrient planning and resource allocation across fields and regions), which helps conserve natural resources and ensure long-term food security~\cite{rose2016decision}. 
At larger scales, these timely estimates improve commodity supply assessments, reduce market uncertainty, 
and improve food-security and disaster-response planning by identifying potential shortfalls early enough for intervention.




Given the importance of this problem, scientists have developed different types of process-based physical models~\cite{stehfest2007simulation,grant1998simulation,zhou2021quantifying} and machine learning (ML) models~\cite{van2020crop,he2023physics,fan2022gnn,liu2024knowledge} for crop modeling. 
Process-based models are designed to 
simulate different components of crop growth based on known physical relationships, 
such as mass and energy conservation laws~\cite{grant2010changes,zhou2021quantifying,jones2003dssat,srinivasan2010swat}. However, these models are necessarily approximations of reality due to incomplete knowledge of certain processes or omission of processes to maintain computational efficiency. 
More recently, ML models have gained significant  attention in environmental monitoring due to  their rapid inference capabilities and ability to automatically learn complex relationships from observational data.

In particular, recurrent neural networks (RNNs), such as long short-term memory (LSTM) and gated recurrent unit (GRU), have demonstrated  superior performance compared to other ML models, such as decision tree-based models, fully connected neural networks, and Transformer-based models, in a wide range of environmental ecosystem modeling tasks~\cite{alibabaei2021crop,liu2025rnns,sharma2020wheat,sun2025x,xiong2024predicting,luo2025learning}. The advantage of RNNs largely stems from their ability to  capture the temporal dependencies inherent in physical systems, where the system states (e.g., soil moisture, crop biomass) evolve in response to external environmental drivers (e.g., weather conditions, fertilization). 
Despite their promise, their effectiveness at large-scale agricultural monitoring remains limited by two major challenges. 
First, although RNNs are effective at capturing short-term temporal dynamics, they are not explicitly designed to model long-term dependencies in crop growth. In practice, crop growth can be influenced by processes operating over multi-year timescales. For example, soil nutrient status may reflect accumulated effects of past farming practices, crop rotations, or residue management~\cite{jernigan2020legacy,congreves2015long}. These long-term factors are difficult to represent using standard RNNs applied to frequent data on a daily scale.

\begin{figure} [!t]
\centering
\subfigure[]{ \label{fig:b}{}
\includegraphics[width=\columnwidth]{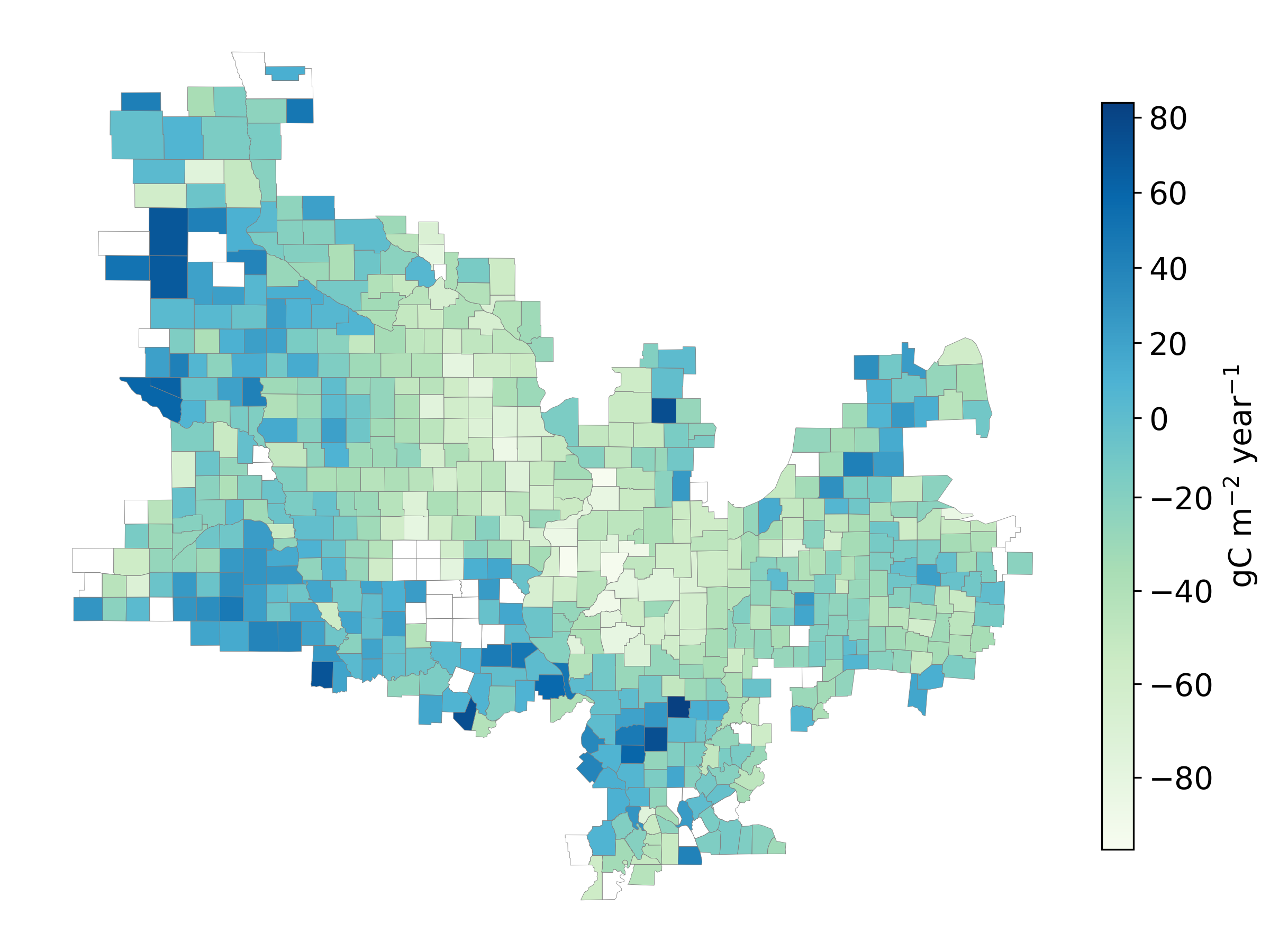}
}\vspace{-.1in}
\subfigure[]{ \label{fig:b}{}
\includegraphics[width=\columnwidth]{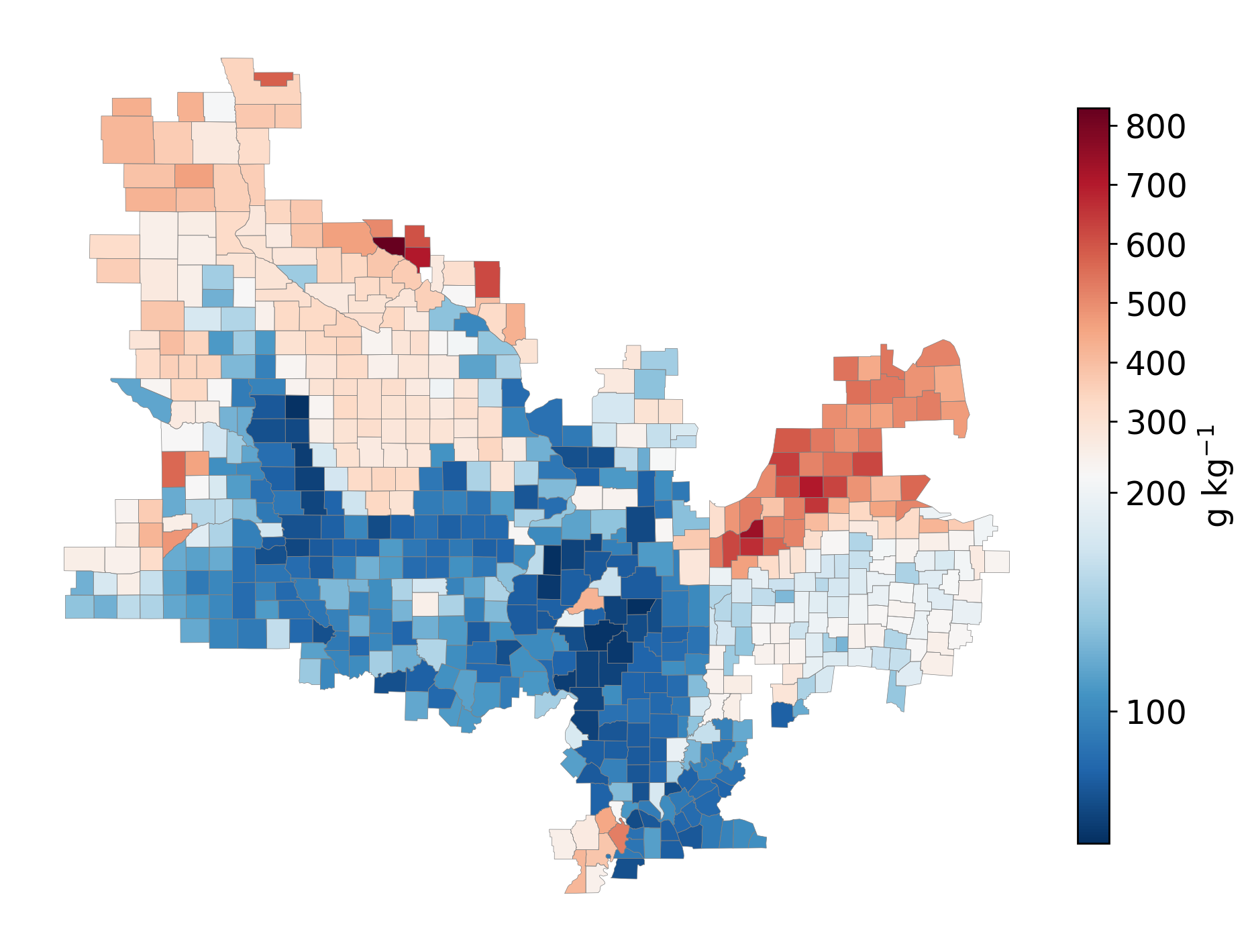}
}
\vspace{-.1in}
\caption{(a) County-level prediction errors (prediction – ground truth) from a single global model trained across all counties. (b) Spatial variability of soil sand content (TSAND) across all counties. 
}
\label{fig:exp}
\end{figure}

Additionally, existing ML methods are typically trained globally across large regions, but their generalizability is often limited by substantial underlying data heterogeneity. In particular, data distributions can vary substantially over space due to differences in environmental conditions and farming practices. 
For example, a global model trained across a large region of 630 counties in the US produces highly uneven prediction errors in a separate testing year (Figure \ref{fig:exp} (a)). This suggests that crop response functions are non-stationary across space, which is mostly because of both environmental differences and variations in management and local farming practices. 
Figure \ref{fig:exp} (b) further illustrates the spatial distribution of soil sand content, a key factor affecting the soil water-holding capacity, nutrient retention, and crop drought sensitivity. Such substantial spatial variation of soil sand content and many other environmental and management factors can affect crop responses even under similar weather conditions.  Temporally, crop growth patterns may also shift across years as a result of technological advancements in farming (e.g., seed quality, irrigation) and changes in environmental conditions (e.g., soil quality). Ignoring such spatial and temporal heterogeneity can lead to unreliable estimates for certain regions or years, which in turn results in uneven or suboptimal decision making (e.g., in resource allocation or insurance assessments).

Prior work has explored spatial-explicit methods that  build separate models for different locations or train a subset of parameters independently for specific spatial regions~\cite{gupta2020towards,xie2021statistically,xie2021spatial}. However, these models often face data scarcity issues, as insufficient data may be collected from each spatial location. Alternatively, transfer learning~\cite{weiss2016survey} and domain generalization~\cite{zhou2022domain,sun2025domain} have been widely studied in the machine learning community to learn generalizable data patterns using a unified model. Nonetheless, directly applying these techniques to agricultural monitoring remains challenging due to the substantial heterogeneity across many different regions.  In particular,  many factors that drive spatial variations (e.g., farming practices, soil conditions) are  unobservable and thus absent from available features, which makes it difficult to directly extract generalizable patterns. 


In this paper, we introduce a novel framework that integrates a new network architecture with a retrieval-based augmentation mechanism.  
The proposed model architecture, Long-term Year-spanning Recurrent Architecture (LYRA), is designed to  capture both daily-scale crop growth dynamics and long-term temporal changes across years. At the short-term scale, LYRA employs the long short-term memory (LSTM) to model daily crop growth patterns, which naturally align with the inherent dynamical processes of agroecosystems, where ecosystem states are recurrently influenced by other fluxes over time. 
At the long-term scale, it utilizes a cross-attention structure to capture long-term dependencies for modeling temporal changes across years. 
To address spatial heterogeneity and data sparsity, we further enhance LYRA with a retrieval-based approach,  inspired by the success of retrieval-augmented generation in natural language processing~\cite{fan2024survey}. This retrieval-based approach aims to adapt LYRA to each location by incorporating relevant historical samples.   
However, due to substantial inter-year variability, the retrieved samples may originate from conditions that differ markedly from those of the target year. As a result, directly using these samples may introduce bias and may not effectively support prediction in the target year.

To mitigate this issue, we introduce a 
Retrieval-augmented Tuning And Refinement (RaTAR) approach, which adjusts the retrieved samples before incorporating them for augmentation. We observe that yearly differences in crop yield, which arise from changes in environmental conditions (e.g., soil properties) and farming technology (e.g., seed quality),  cannot be explained by available input features. To account for these unobserved variations, we first estimate pairwise yearly differences by evaluating models trained on one year against data in other years. The estimated yearly differences are then used to correct biases in retrieved samples, which enables more effective augmentation.

We evaluate the proposed method in the context of crop yield prediction over 630 counties spanning the U.S. Corn Belt States.  The results demonstrate the effectiveness of  the LYRA model and the RaTAR learning strategy. The proposed method opens new opportunities for a broad range of Earth science problems where sparse data are collected from large spatial regions with data heterogeneity. 




\section{Related Work}

\subsection{Spatio-temporal crop modeling} 
Precise accounting of crop-related variables, such as crop yield, biomass, carbon and nitrogen emissions, is important  for ensuring food security, which becomes increasingly important given the growing population. 
Researchers have been extensively exploring data-driven approaches for modeling crops. Crop growth typically exhibits temporal patterns in response to external drivers (e.g., weather and  management practices) and  spatial dependencies given the continuity of environmental conditions. To explicitly capture such relationships, prior works have explored crop modeling using recurrent temporal models (e.g., LSTM or GRU)~\cite{sharma2020wheat,bhimavarapu2023improved,he2023physics,wang2022winter}, Transformer-based models~\cite{lin2023mmst,osibo2024tcnt,bi2023transformer}, and graph neural networks~\cite{fan2022gnn,sarkar2024crop,jiang2020cnn}. 
More recently, researchers have also started developing hybrid modeling by integrating physical and biochemical knowledge into data-driven models~\cite{liu2024knowledge,liu2022kgml,he2023physics}.

There are two main  challenges faced by data-driven crop modeling. First, crop growth is affected by both short-term weather and long-term temporal factors. Temporal convolutional networks (TCN) have been commonly used to capture such multi-scale temporal patterns in crop modeling~\cite{gong2021deep,gavahi2021deepyield,mohan2023temporal}. In contrast, we introduce a complementary design that explicitly decomposes temporal dynamics: a GRU module is used to represent fine-grained, daily-scale variations, while an attention mechanism captures longer-term dependencies across yearly growth cycles. This architecture is more closely aligned with known crop phenological patterns and the distinct temporal regimes governing yield formation. The second challenge lies in the spatial variability of crop patterns  due to different weather conditions and farming practices (e.g., fertilization, seeding, irrigation) in different regions. Hence, a global model trained over large regions may not fit each specific location~\cite{karpatne2018machine,goodchild2021replication}. Prior work has investigated explicit tuning of model parameters towards each location~\cite{gupta2020towards,xie2021statistically,xie2021spatial}
and meta-learning approaches to facilitate region-specific model adaptation~\cite{liu2023task,tseng2021learning,reuss2025eurocropsml}. However, these methods could suffer from limited data available for local tuning.

\subsection{Retrieval-augmented generation in scientific problems}

Effective task (location)-specific adaptation can help better align the learned model with the characteristics of the target task.
Separate modeling of each individual task generally faces data scarcity because of their restricted spatial and temporal scope and the practical limitations of in-situ measurement systems. One promising direction is to leverage retrieval-augmented generation (RAG), where external information can be retrieved and incorporated into the inference process. RAG has demonstrated its capabilities in knowledge-intensive natural language processing tasks, particularly in open-domain question answering where retrieval and synthesis from large relevant text corpora are essential~\cite{gupta2024comprehensive,hu2024rag}. However, applying RAG in scientific problems requires addressing the challenges in  effectively retrieving and referencing  external knowledge, which are less explored by existing RAG methods~\cite{karpatne2025ai,yu2025rag}. 
In particular, retrieval needs to  prioritize data points whose scientific relevance is consistent with the underlying physical processes. This often requires defining new relevance measures among different types of data representations, such as spatial data~\cite{yu2025spatial}, time series~\cite{luo2025learning,ning2025ts}, and knowledge graphs~\cite{jiang2025retrieval}.

Moreover, due to measurement limitations, data collected from scientific systems are often incomplete and insufficient to capture all system characteristics. This incompleteness can reduce the reliability of retrieval, as data embeddings and relevance scores are derived only from the available data. 
Knowledge refinement and editing~\cite{wang2024knowledge} is a promising direction to further enhance the reliability of the retrieved data. Specifically, retrieved information can be further refined, e.g., through filtering~\cite{yan2024corrective,wang2023learning}, completion~\cite{narita2025chunk}, and retriever adjustment~\cite{asl2025fair,anantha2024context}, 
to fit the target scientific system. Such refinement not only improves data quality but also enhances the coherence of the knowledge supporting the reasoning within RAG. 



\section{Problem Definition}


Our objective is to learn a model $\mathcal{F}_\theta$ that estimates the annual county-level corn yield  $y_i^k$ for each county $i$ in year $k$, based on its  daily input features $\mathbf{x}_{i}^k = \{\mathbf{x}_{i,t}^k\}_{t=1:T}$   ($T$=365) within that year. Daily features $\mathbf{x}_{i,t}^k$ include weather drivers (e.g., precipitation, solar radiation) and  soil and crop properties. The feature values are obtained as the average across a set of randomly sampled farm locations in each county.  More details are provided in Section~\ref{sec:dataset}. 

This year targets the current-year prediction problem, which aims to predict crop yield for a specific year assuming this year's input features and historical yield records are available. This  setting is practically important for informing policy and management decisions for upcoming seasons (e.g., irrigation and nutrient planning and resource allocation across fields and regions). The predictions can also help provide timely information to support food supply assessment and food-security planning. 
Specifically, we have access to crop yield labels from 630 counties in the US for previous years $k=1,...,K$, which serve as the training data. During testing, we apply the model to the next year $K+1$, where only the input features are available and the yield label is withheld for evaluation.


\begin{figure*}[t]
  \centering
  \includegraphics[width=\linewidth]{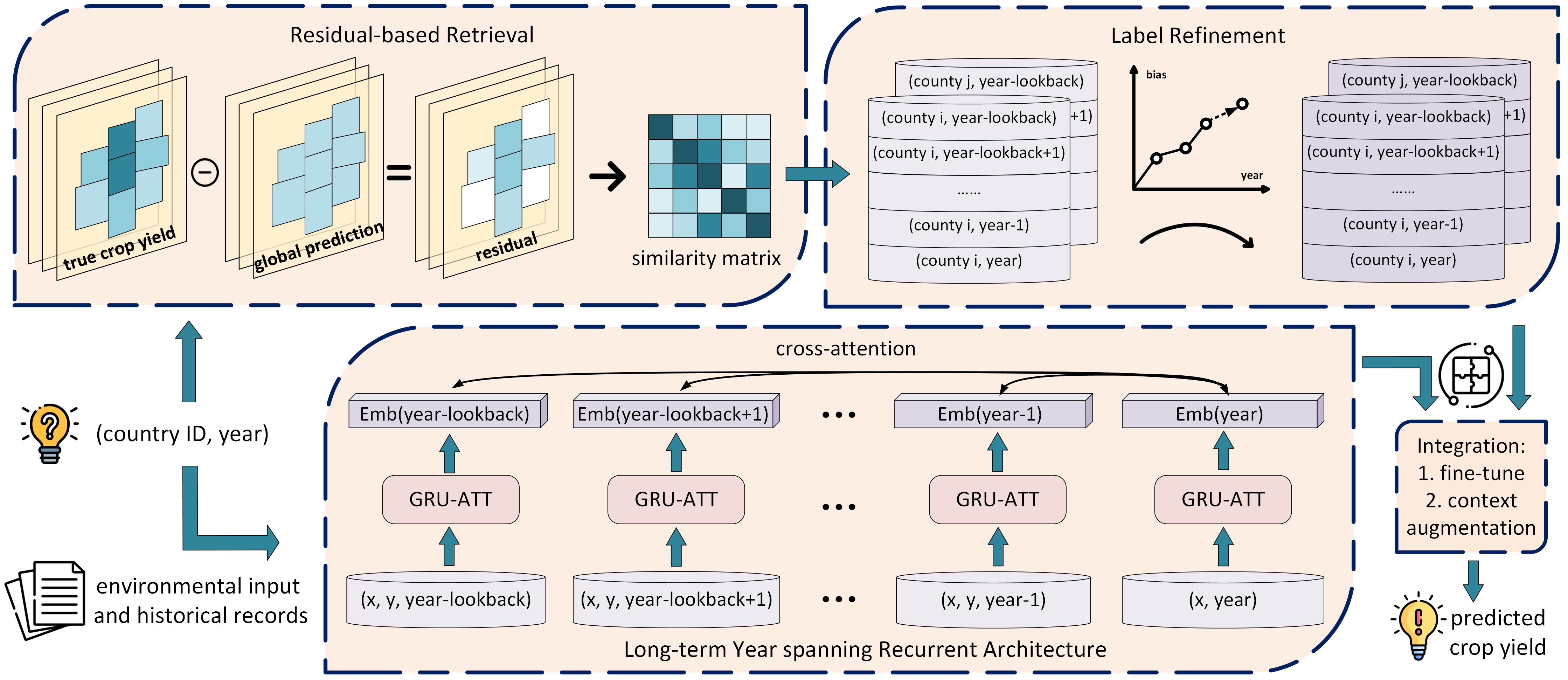}
  \vspace{-0.1in}
  \caption{The overall framework of LYRA-RaTAR.}
  \vspace{-0.05in}
  \label{fig:model}
\end{figure*}

\section{Method}

Our proposed method consists of two key components, a backbone model  (LYRA) and a retrieval-based adaptation strategy (RaTAR), as illustrated in Figure~\ref{fig:model}.  The LYRA model is trained on the entire training dataset, and is designed to capture both short-term and long-term temporal dynamics in crop growth. 
During prediction in the testing year, we enhance LYRA using the retrieval-based approach RaTAR. Specifically, for each county, 
we retrieve historical data samples from other counties showing similar crop behaviors, refine these samples, and incorporate them to improve the prediction. In the following, we provide details for these two components.

\subsection{Long-term Year spanning Recurrent Architecture}

Environmental ecosystems are inherently dynamical physical systems in which system states (e.g., soil water conditions, crop biomass, nutrient availability) evolve over time in response to dynamic external forcings, which are often referred to as input drivers to the ecosystem (e.g., weather variables, management practices). During the system evolution, some system measurements can be collected, either for certain flux variables (e.g., latent heat flux, carbon dioxide) using instruments such as eddy covariance towers,  or for accumulated quantities (e.g., crop yield) through large-scale field surveys. 
RNNs (e.g., LSTM and GRU) have been widely applied to environmental ecosystem modeling across many domains and have generally  achieved better performance than many other ML models~\cite{alibabaei2021crop,liu2025rnns,sharma2020wheat,sun2025x,xiong2024predicting,luo2025learning,he2023physics}.  This is because the recurrent  network structure used in these models can naturally capture the temporal evolution of system states in a manner analogous to the dynamics in physical systems. 
However, RNNs remain limited in modeling long-term dependencies that characterize certain slow-evolving processes, especially when operating on high-frequency data (e.g., on a daily scale). In agricultural ecosystems, for example, crop growth depends on soil properties that are often shaped by the cumulative impacts of farming practices over multiple past years~\cite{jernigan2020legacy,congreves2015long}.

Our proposed LYRA leverages the capacity of recurrent neural networks (e.g., LSTM and GRU) to embed the short-term temporal dynamics of crop growth. In addition, it  captures long-term dependencies across years, which enables the model to account explicitly for the influence of conditions in previous years. 

\textbf{Generation of yearly embedding: } For each year $k$, LYRA employs GRU layers over the daily input sequence within this year, as $\{\mathbf{h}^k_{i,t}\}_{t=1}^T = \text{GRU}(\{\mathbf{x}_{i,t}^k\}_{t=1}^T)$.   
Next, we aim to combine the obtained daily embeddings $\mathbf{h}^k_{i,t}$ from different dates $t=\{1,...,T\}$ to create a yearly embedding for year $k$ through an attention layer. The attention weights are estimated from the hidden representation through a multi-layer perceptron (MLP) and normalized across all time steps $t = 1, \dots, T$ using a softmax function, as follows:  
\begin{equation}
\begin{aligned}
\alpha^k_{i,t} &= \text{MLP}(\mathbf{h}^k_{i,t}), \\
\bar{\alpha}^k_{i,t} &= \text{softmax}(\alpha^k_{i,t}).
\end{aligned}
\label{eq:att_weights}
\end{equation}



We then aggregate the daily embeddings based on the attention weights and also 
incorporate the yield label $ y_i^k$ and year index $k$ to create the yearly embedding $\mathbf{z}_i^k$, as follows:
\begin{equation}
{\mathbf{z}}_i^k = \text{MLP}([\sum_t \bar{\alpha}^k_{i,t} \mathbf{h}^k_{i,t}, y_i^k, \text{Embed}(k)]),
\label{eq:yearly_embedding}
\end{equation}
where  the function $\text{Embed}(\cdot)$ directly maps each year index to a trainable continuous embedding. 
During testing, as we do not have the yield label, we replace the $y_i^k$ in Eq.~\ref{eq:yearly_embedding}
by the predicted yield of a GRU model trained over the entire training set. 


This GRU-based structure enables capturing short-term dynamics of crop growth, such as the influence of weather changes (e.g., heavy rainfall)  over a short period of time. The attention mechanism described in Eqs.~\ref{eq:att_weights} and ~\ref{eq:yearly_embedding} further enables the model to  automatically identify the most informative time periods in a year that are most relevant for yield prediction, such as the key stages of the growing season. However, crop growth could also be affected by many other factors over a longer time horizon, e.g., soil conditions in the current year may be shaped by farming practices in previous years or by how crop residues were managed. 
This motivates capturing long-term dependencies across years.

\textbf{Cross-year attention mechanism: }For predicting the testing year $K+1$, we construct a look-back window consisting of previous $w$ years, i.e., from year $K-w+1$ to $K$, to serve as  contextual information for predicting the yield in the testing year $K+1$. For each county $i$, we generate the yearly embeddings $\mathbf{z}_i^k$ (following Eq.~\ref{eq:yearly_embedding}) for the testing year and each year in the look-back window. 
Then a cross-year attention is applied between the testing year and  each year in the look-back window  using their yearly embeddings. The attention weights are normalized over the historical years in the look-back window using a softmax function, as follows:
\begin{equation}
\begin{aligned}
\beta &= \text{softmax}(\{\mathbf{z}_i^{k} \cdot \mathbf{z}_i^{K+1}\}_{k=K-w+1:K}). 
\end{aligned}
\end{equation}

Then we aggregate the influence from previous years based on their attention weights, and add such influence to the original yearly embedding of the testing year: 
\begin{equation}
\begin{aligned}
\tilde{\mathbf{z}}_i^{K+1} &= \mathbf{z}_i^{K+1} + \sum_{k=K-w+1:K} \beta_{k} \mathbf{z}_i^{k}.
\end{aligned}
\end{equation}

Finally, we generate the prediction $\hat{y}_i^{K+1}$ for county $i$ in the testing year $K+1$ using an MLP, as $\hat{y}_i^{K+1} = \text{MLP}(\tilde{\mathbf{z}}_i^{K+1})$. 
LYRA is trained globally on the entire training set $\mathcal{D}_\text{tr}$, i.e., by minimizing the mean squared loss using data from  all the counties in the training years. 


\subsection{Retrieval-augmented Tuning And Refinement (RaTAR)}
\label{sec:method_retr}

Although the backbone model LYRA can capture multi-scale temporal dynamics, it remains limited for handling spatially heterogeneous data across different counties. Intuitively, we can adapt the global LYRA model separately to each individual county through fine-tuning. However, this often leads to degraded performance due to limited data in each county (i.e., labeled data from a specific county in previous $K$ years). We address this issue using a new retrieval-based approach RaTAR, which consists of three steps: retrieval, refinement, and integration.

\textbf{Step 1. Residual-based retrieval: } 
A major source of spatial heterogeneity is that certain influential factors, such as soil properties or specific farming practices, are not directly observable or measurable,  yet these hidden variables can substantially alter the relationship between input drivers and crop yield. Ideally, we would like to retrieve historical data from locations that share similar unobserved characteristics with the target location to improve prediction accuracy.  
Hence, to facilitate the learning of location-specific patterns, we propose a residual-based approach that emphasizes the components of temporal data variations not explained by a global model that learns the input-output relation. 

In particular, we construct a GRU-based global model $f(\cdot)$ from all the training data. Then we apply the global model to each county and measure the residual, i.e., the difference $r_i^k = y_i^k-f(\mathbf{x}_i^k)$ for each training year $k=1,2,...,K$, which quantifies the model's unexplained error for county $i$ in year $k$. Collecting these values forms a residual vector for each county $\mathbf{r}_i = [r_i^1, r_i^2, ..., r_i^K]$. 
Next, we measure the 
similarity between two counties $i$ and $j$  using the cosine similarity, as  $\text{sim}(i,j) = \text{cosine}(\mathbf{r}_i-\text{mean}(\mathbf{r}_i), \mathbf{r}_j - \text{mean}(\mathbf{r}_j))$. 
For each target county $i$, we retrieve  counties whose similarity to county $i$ is greater than a threshold (set as 0.9 in our tests). Then we collect the data samples from these counties over the most recent five years during the augmentation process to enhance the learning of location-specific dynamics.

\textbf{Step 2. Refining the retrieved samples: }
Due to the difference in crop yield across years, we aim to refine the crop yield labels of the retrieved samples before incorporating them into the prediction of the testing year. Specifically, for every pair of years $s$ (source) and $k$ (target), we train a model $g_s(\cdot)$ that directly predicts labels from $\mathbf{z}^s$ to $y^s$ on the source data, and then apply it to the embedding $\mathbf{z}_i^k$ for each location $i$   in the target year $k$ and measure the model bias $b^{sk}_i$ as $y_i^k-g_s(\mathbf{z}_i^k)$. 
Once we gather these values, we create a cross-year bias matrix $\mathbf{B}_i = \{b_i^{sk}\}_{s=1:K, k=1:K}$. To refine a retrieved sample from year $s$ towards the testing year $K+1$, we perform a linear 
extrapolation on $B_i[s,:]$ to estimate the bias $\hat{b}_i^{s,K+1}$. Finally, we refine the label of retrieved $y_i^s$ as $\tilde{y}_i^{s} \sim \mathcal{N}(y_i^s + \hat{b}_i^{s,K+1}, \sigma^2)$, 
where the Gaussian sampling  with a pre-defined variance $\sigma^2$ helps  increase the diversity of the refined samples.

\textbf{Step 3. Integration of retrieved samples:  }
For each  target location $i$ as the query, we  use the set of retrieved samples and their refined labels 
$\mathcal{R}_i = \{\mathbf{x}_i^s, \tilde{y}_i^s\}$
to enhance the prediction in the testing year. Specifically, we consider two approaches: 
\begin{itemize}
\item \textit{Fine-tuning}: we use the retrieved samples $\mathcal{R}_i$  to fine-tune the LYRA model.
\item \textit{Context augmentation}: We add the retrieved samples $\mathcal{R}_i$ to the context (i.e., the look-back window) of the LYRA model.
\end{itemize}

Our experiments (Section~\ref{sec:exp-retr}) show that both approaches can effectively improve the predictive accuracy.

\section{Experiments}

\subsection{Dataset}
\label{sec:dataset}

We use the corn yield data from 630 counties  in the midwestern United States from the years 2000-2020 provided by USDA National Agricultural Statistics Service\footnote{\url{https://quickstats.nass.usda.gov/}}. 
These counties span the major U.S. Corn Belt states (e.g., Iowa, Illinois, Indiana, Minnesota, Missouri, Nebraska), covering diverse environmental and management conditions. 
They were selected using the USDA Cropland Data Layers  and Corn-Soy Data Layer, retaining counties with a combined corn and soybean fraction greater than  0.2~\cite{wang2020mapping}.  
The corn yield data (in gC m$^{-2}$) are available for each county in each year. 
The input features have 19 dimensions, including NLDAS-2 climate data~\cite{xia2012continental}, 0-30 cm gSSURGO soil properties\footnote{\url{https://gdg.sc.egov.usda.gov/}}, crop type information,
the 250 m Soil Adjusted Near-Infrared Reflectance of vegetation (SANIRv) based daily GPP product regridded to county level~\cite{jiang2021daily}, and calendar year. 
Daily climate and GPP sequences are used directly as model inputs for each year, while soil variables are static. All continuous variables are standardized using z-score normalization. Comprehensive dataset descriptions are provided in Appendix~\ref{appendix:dataset}.


\begin{table*}[!t]
\newcommand{\tabincell}[2]{\begin{tabular}{@{}#1@{}}#2\end{tabular}}
\centering
\caption{The prediction root mean squared errors (RMSE) by different approaches in each testing year. We report the mean and standard deviation over 3 runs with random model initialization. The best results are highlighted in bold. }
\vspace{0.05in}
\begin{tabular}{l|ccccc}
\hline
Method& 2016 & 2017  & 2018& 2019& 2020\\ \hline 
LSTM & 36.37$\pm$0.90 & 35.02$\pm$1.18 & 38.82$\pm$3.72 & 32.72$\pm$3.59 & 31.40$\pm$1.41 \\
TCN & 30.93$\pm$6.09 & 35.44$\pm$2.37 & 36.71$\pm$3.30 & 30.81$\pm$0.85 & 30.69$\pm$0.80 \\
Transformer & 43.10$\pm$1.53 & 48.79$\pm$3.55 & 53.80$\pm$6.44 & 45.20$\pm$1.71 & 39.55$\pm$2.61 \\
Pyraformer & 57.53$\pm$0.56 & 56.12$\pm$1.72 & 68.32$\pm$7.61 & 50.43$\pm$0.04 & 41.90$\pm$1.14 \\
iTransformer & 60.94$\pm$1.46 & 58.87$\pm$3.93 & 65.05$\pm$6.40 & 50.53$\pm$0.10 & 42.93$\pm$1.61 \\
TS-RAG & 45.54$\pm$6.25 & 41.48$\pm$2.09 & 46.38$\pm$3.37 & 36.80$\pm$3.12 & 34.39$\pm$1.50 \\
LYRA-RaTAR & \textbf{28.35$\pm$1.50} & \textbf{30.13$\pm$1.43} & \textbf{34.87$\pm$2.03} & \textbf{27.32$\pm$1.92} & \textbf{27.79$\pm$1.58}\\ \hline 
\end{tabular}
\label{tab:rmse}
\end{table*}

\subsection{Experimental settings}
We consider the yield prediction for the next unseen year. Hence, we treat the labeled data from previous years as training data and aim to predict the corn yield for the next year. In particular, we treat each year from 2016 to 2020  as a separate testing year. For each testing year, all labeled data from years prior to the testing year are used for training.  
We compare the proposed method with a set of baselines described below. All baselines operate at the daily time scale, producing 365 daily outputs. To generate the annual yield prediction, we apply mean pooling over the daily embeddings followed by a multi-layer perceptron. 
\begin{itemize}[leftmargin=2em]
\item RNN-based models: (a) LSTM: A recurrent neural network that captures temporal dependencies through gated memory units. Widely used in agricultural and environmental forecasting~\cite{LSTM}.  (b) TS-RAG (LSTM backbone): A retrieval-augmented model that retrieves semantically similar historical sequences under comparable environmental conditions~\cite{ning2025ts}.
\item CNN-based model: (a) TCN: A temporal convolutional network that models long-range temporal patterns using dilated causal convolutions~\cite{liu2019time}. 
\item Transformer-based models: (a) Transformer: A self-attention encoder-only architecture that captures global temporal dependencies without recurrence~\cite{vaswani2017attention}. (b) Pyraformer: A hierarchical, multi-resolution Transformer that reduces attention complexity through pyramid-shaped attention patterns, enabling efficient long-sequence modeling~\cite{liu2022pyraformer}. (c) iTransformer: An instance-centric Transformer that treats each variable as a token, enabling cross-variable interaction compared to standard time-position encoding~\cite{liu2023itransformer}.
\end{itemize}







\subsection{Predictive performance}

Table~\ref{tab:rmse} shows the performance of LYRA-RaTAR and baseline methods in each testing year. 
Here the proposed LYRA-RaTAR incorporates the retrieved samples using the fine-tuning approach. We compare it with the context augmentation approach in Section~\ref{sec:exp-retr}. 
The results in Table~\ref{tab:rmse} show that the proposed method significantly outperforms all the other methods in terms of predictive accuracy. 
LYRA-RaTAR’s performance is also highly reliable, consistently exhibiting low standard deviation among all testing years. 
Critically, all methods perform worse in 2018, which is partially due to the distinct drought patterns in that year~\cite{drought_2018}. Such performance degradation verifies the negative impact of inter-annual climate variability on the performance, 
yet LYRA-RaTAR maintains the highest relative accuracy and stability even when confronted with these severe environmental shifts.

Analyzing the architectural differences, LSTM generally performs better than Transformer-based models. This suggests that the recurrent structure of the LSTM is better suited for capturing the temporal evolution of crop growth, as it more closely resembles a dynamical system whose internal state evolves in response to external environmental forcings. The convolutional TCN, which also emphasizes local, time-ordered dependencies, aligns with this finding.  Furthermore, LYRA-RaTAR’s  performance advantage over the retrieval-augmented TS-RAG model validates the strength of its specialized augmentation approach. Crucially, TS-RAG, which relies on retrieval solely based on environmental similarity (using input features), performs significantly worse than even the baseline LSTM. This pronounced failure suggests that solely relying on environmental information is insufficient due to the non-stationary nature of the input-output relationship.



\begin{figure*}[t]
  \centering
  \includegraphics[width=0.88\linewidth]{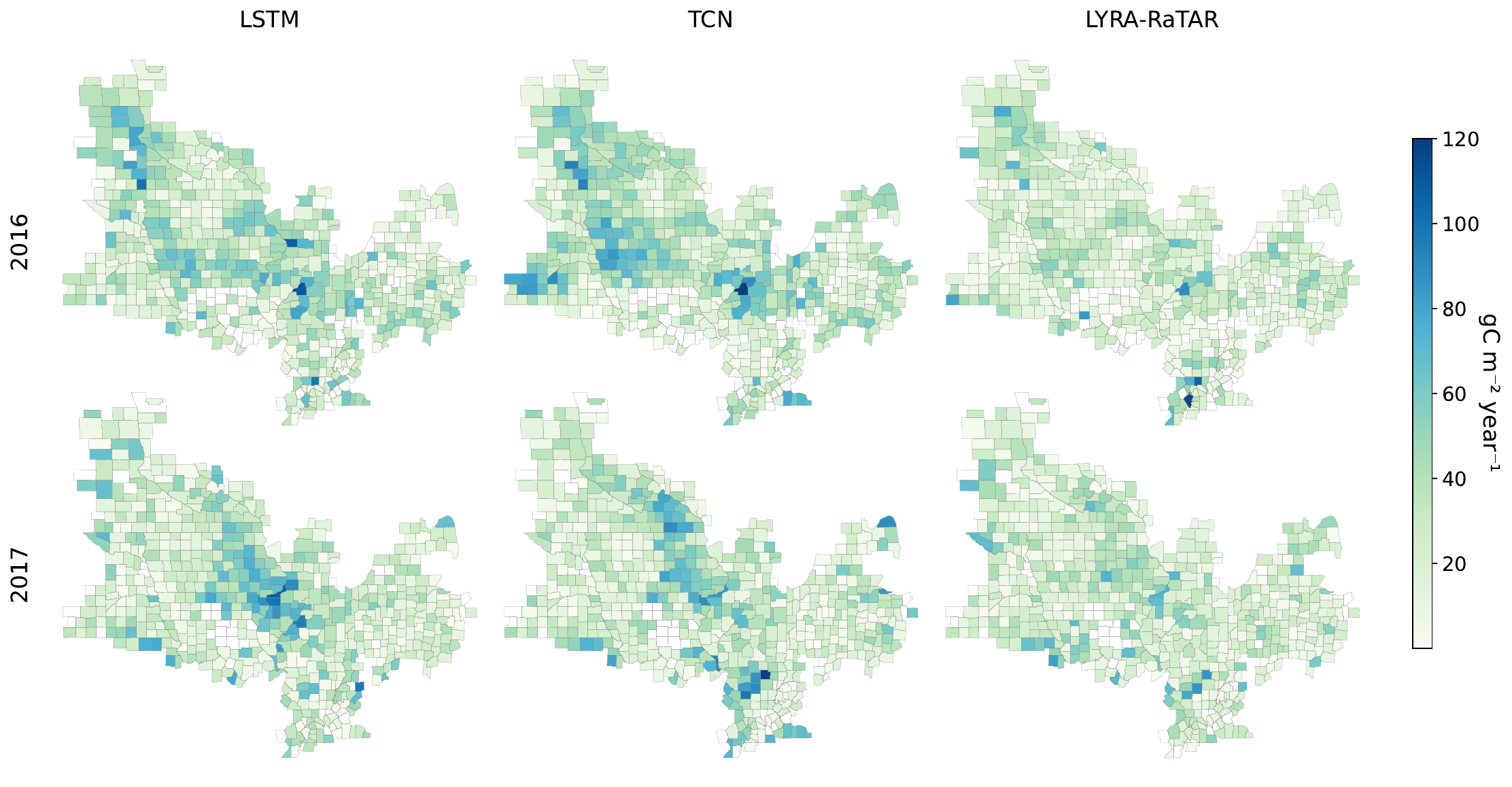}
  \caption{Spatial patterns of prediction errors for LSTM, TCN, and LYRA-RaTAR evaluated on the 2016 and 2017 test years.}
  \label{fig:error16}
\end{figure*}

In Figure~\ref{fig:error16}, we show the spatial distribution of prediction errors produced by three different methods across all counties. The global baselines (LSTM and TCN) exhibit large errors in many regions, suggesting limited generalization under substantial spatial heterogeneity. In contrast, LYRA-RaTAR reduces several of these high-error regions by leveraging retrieved county-specific historical patterns, yielding more spatially consistent predictions. While some counties remain challenging, the improvement highlights the value of incorporating retrieval to adapt models to spatially variable agricultural ecosystems.

\subsection{Ablation studies}

To evaluate the contribution of each component in the proposed method, we conduct ablation studies and report the performance of several variants of RaTAR in Table~\ref{tab:ablation}. These variants include: `w/o refine', which  directly uses the retrieved samples without refinement for location-specific adaptation of LYRA; `LYRA',  which retains the same network architecture but does not perform  retrieval; and `GRU-ATT', which removes the cross-year dependency modeling from LYRA. 

It can be seen that the largest performance gain comes from introducing the cross-year attention in LYRA. 
This demonstrates the importance of modeling long-term temporal dependencies, which complements the short-term modeling by RNNs for crop yield prediction. 
The retrieval and refinement modules provide additional improvements, with particularly notable gains  in years 2018 to 2020. This is partly attributable to the higher spatial variability  due to more frequent extreme events (e.g., floods and drought) in certain regions of the US in these years compared to 2016-2017~\cite{extreme_20182020}. The retrieval-augmented prediction helps better capture distinct patterns for each location compared to a global model applied to all the locations.   

\begin{table}[!t]
\newcommand{\tabincell}[2]{\begin{tabular}{@{}#1@{}}#2\end{tabular}}
\vspace{0.1in}
\centering
\caption{Performance (in RMSE) for different variants of the proposed method.  }
\begin{tabular}{l|ccccc}
\hline
\textbf{Method}& 2016 & 2017  & 2018& 2019& 2020\\ \hline 
LYRA-RaTAR & 28.46 & 30.04 & 34.92 & 27.21 & 27.83\\ 
w/o refine & 29.22 & 30.18 &  35.72  &  27.67  &  28.46\\ 
LYRA & 29.97 & 31.15 &  37.37  &  30.84 &   29.88\\ 
GRU-ATT & 34.38 & 37.25 &  39.97 &   34.52 &   31.90\\ 
\hline 
\end{tabular}
\vspace{0.05in}
\label{tab:ablation}
\end{table}

\subsection{Analysis  of retrieval methods}
\label{sec:exp-retr}


Here we investigate the impact of different strategies for integrating retrieved samples, as well as the effects of various relevance measures used during retrieval. Finally, we analyze the sensitivity of model performance to the number of retrieved samples.

\begin{figure}[t]
  \centering
  \includegraphics[width=0.75\linewidth]{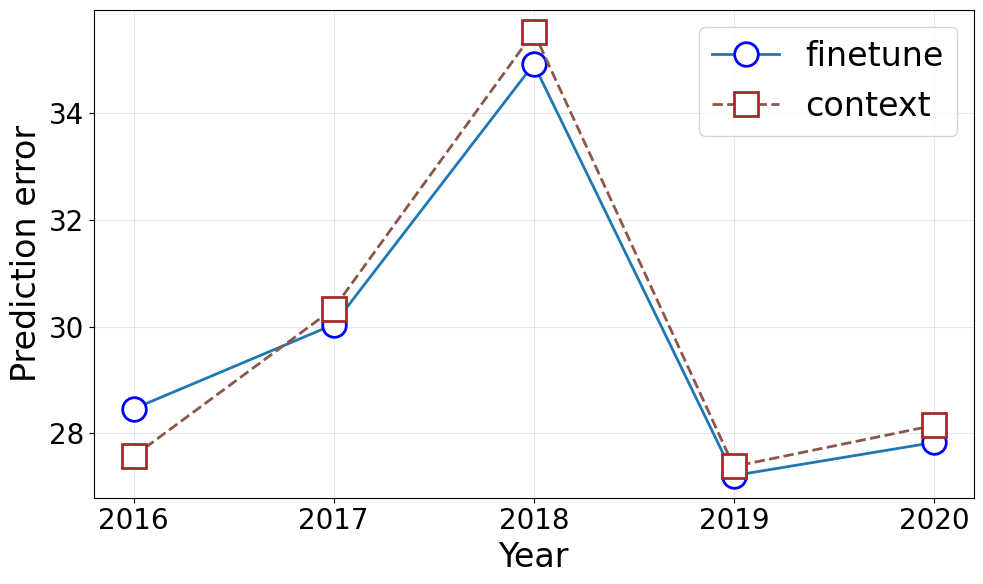}
  \vspace{-0.05in}
  \caption{Performance (in RMSE) of different sample integration methods (fine-tuning vs. context augmentation) in testing years from 2016 to 2020.}
  \label{fig:int_method}
\end{figure}

We test two different strategies for integrating retrieved samples, either using them for location-specific fine-tuning  or including them in the context (i.e., look-back window), as described in Section~\ref{sec:method_retr}.  
Figure~\ref{fig:int_method} shows the performance of these two strategies across all the testing years. We can see that they achieve similar performance in all the testing years.

\begin{figure}[t]
  \centering
  \includegraphics[width=0.7\linewidth]{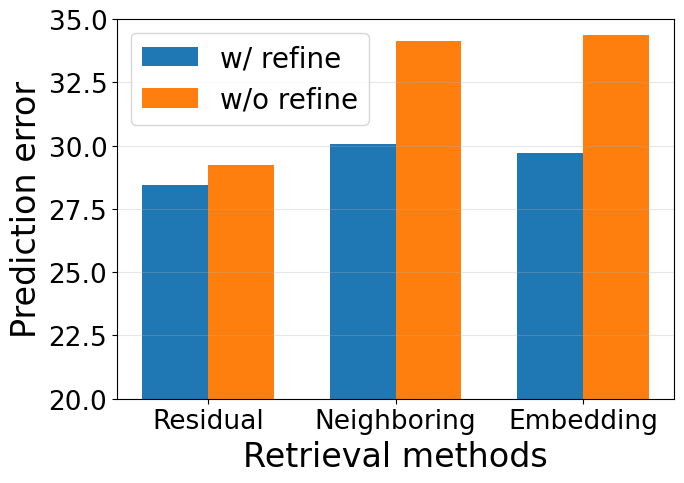}
  \vspace{-0.05in}
  \caption{Comparison of three approaches for retrieving relevant samples: residual-based retrieval, neighboring-based retrieval, and embedding-based retrieval. For each approach, we show the results for the fine-tuned LYRA model using both the original retrieved samples (w/o refine) and the refined retrieved samples (w/refine) in testing year 2016. }
  \label{fig:retr_method}
\end{figure}

We also investigate the effect of different relevance measures in retrieval. Besides the residual-based approach used in RaTAR (described in Section~\ref{sec:method_retr}), we also consider two other ways of identifying relevant samples: (1) selecting data from geographically adjacent counties (Neighboring) and (2) selecting data from counties with similar LYRA embeddings (Embedding).  For both strategies, following the same setting with RaTAR, we retrieve the data samples from the previous five years from the counties selected by Neighboring or Embedding. Figure~\ref{fig:retr_method} shows the predictive errors using the retrieved samples based on each relevance measure. The proposed residual-based relevance yields a better performance than both Neighboring and Embedding. Unlike the other two methods, RaTAR retrieves samples based on similarity in temporal residuals, allowing it to capture meaningful environmental and management shifts specific to the target location.



\begin{figure}[t]
  \centering
  \includegraphics[width=0.8\linewidth]{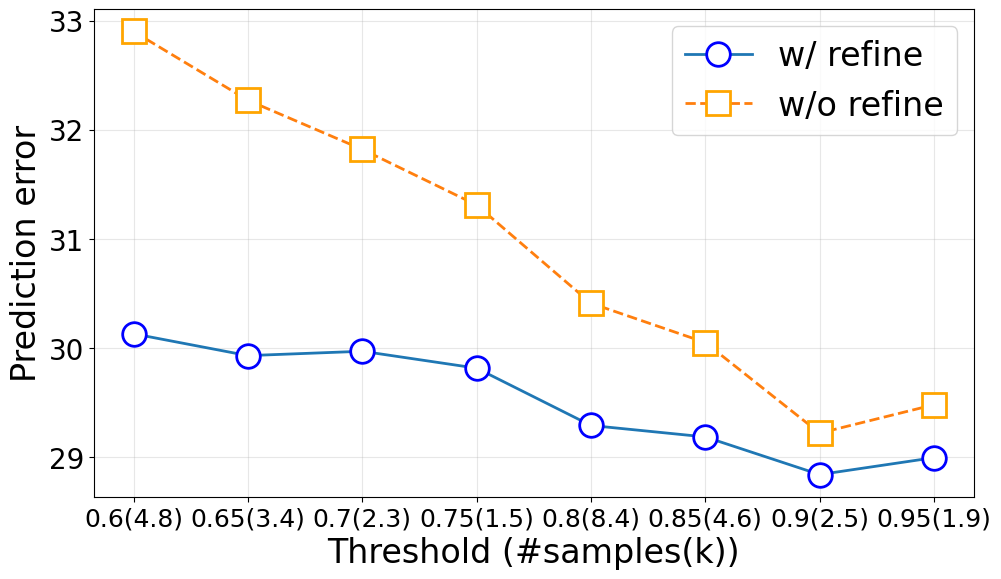}
  \vspace{-0.05in}
  \caption{Performance variation of the proposed method in the testing year 2016  with different numbers of retrieved samples, shown for two model versions: with refinement (w/ refine) and without  refinement (w/o refine). }
  \vspace{0.2cm}
  \label{fig:num_samples}
\end{figure}

We further study how the number of retrieved samples  affects model performance (Figure~\ref{fig:num_samples}). Here x-axis shows the sum of the total number of counties retrieved across different target counties. 
When more samples are retrieved by relaxing the relevance threshold, we observe a performance degradation. This is largely because a greater proportion of irrelevant samples are included. Conversely, increasing the threshold from 0.9 to 0.95 also harms performance, as the resulting retrieved samples become too small to provide effective augmentation. Comparing the variants with and without refinement shows that RaTAR’s refinement step can partially offset the negative effects of retrieving an excessive number of samples.

\subsection{Analysis of LYRA}

\begin{figure}[t]
  \centering
  \includegraphics[width=0.78\linewidth]{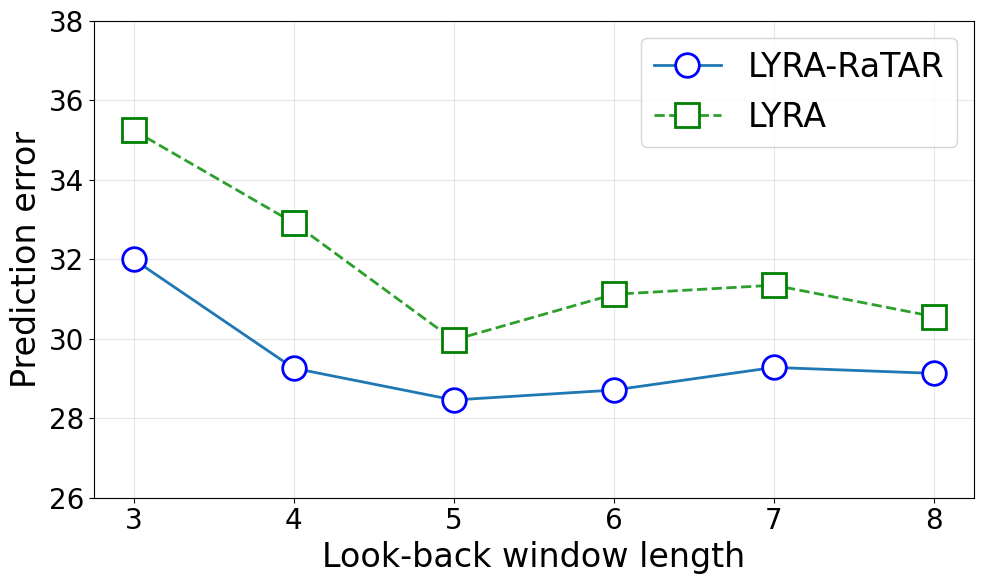}
  \vspace{-0.05in}
  \caption{Performance of the proposed  LYRA-RaTAR  and LYRA (i.e., the same backbone model w/o  retrieval) in 2016 with different numbers of years in the look-back window.}
  \vspace{0.05in}
  \label{fig:lb_size}
\end{figure}

We now analyze the impact of the cross-year attention mechanism in LYRA. As shown in Figure~\ref{fig:lb_size}, the performance exhibits similar trends with and without retrieval (RaTAR vs. LYRA) as we vary the number of historical years included in the context. When the look-back window is short, performance declines due to insufficient historical information. As more historical years are incorporated, the model benefits from the richer long-term context, leading to improved predictive accuracy. The model performance slightly declines when the look-back window exceeds five years, which reveals that additional distant years may introduce more noise and less relevant information.


\begin{figure} [!t]
\centering
\subfigure[]{ \label{fig:b}{}
\includegraphics[width=0.4\columnwidth]{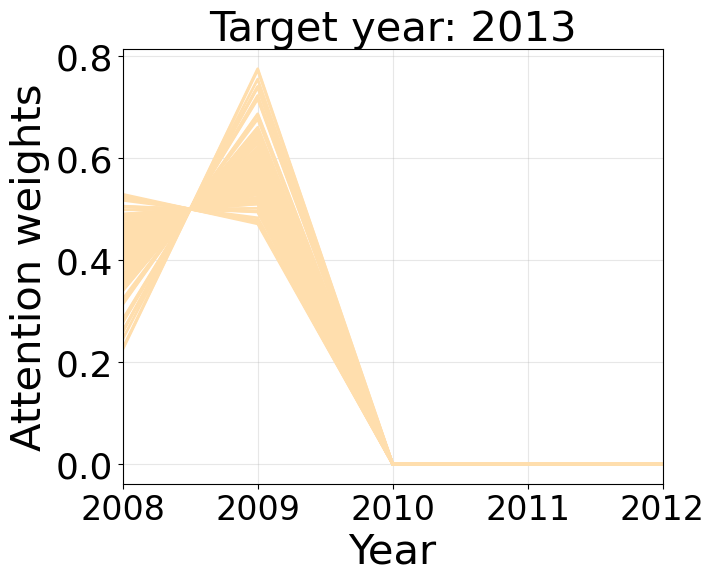}
}
\subfigure[]{ \label{fig:b}{}
\includegraphics[width=0.4\columnwidth]{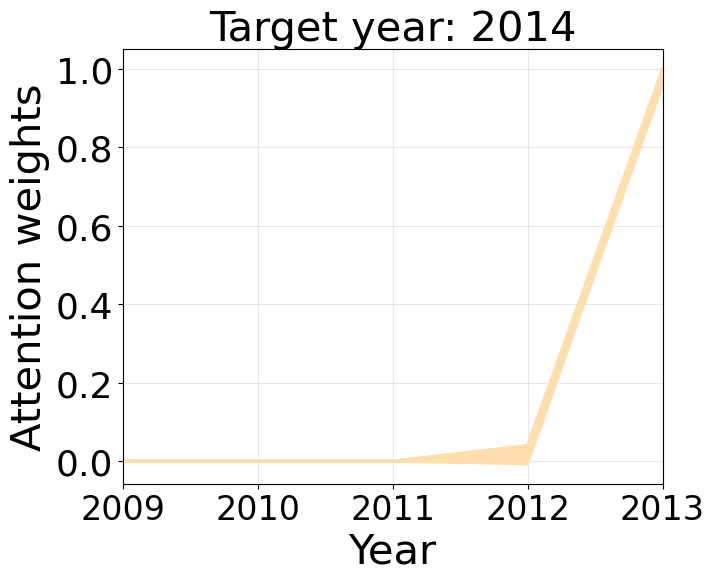}
}\vspace{-0.1in}
\subfigure[]{ \label{fig:b}{}
\includegraphics[width=0.4\columnwidth]{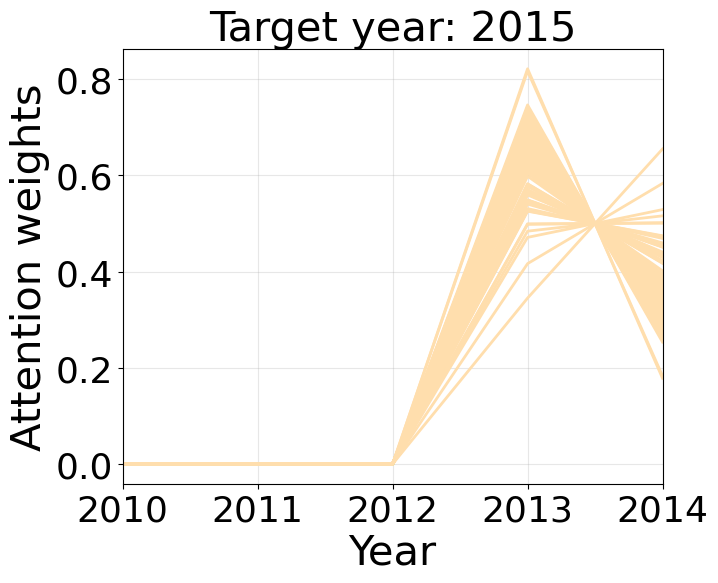}
}
\subfigure[]{ \label{fig:b}{}
\includegraphics[width=0.4\columnwidth]{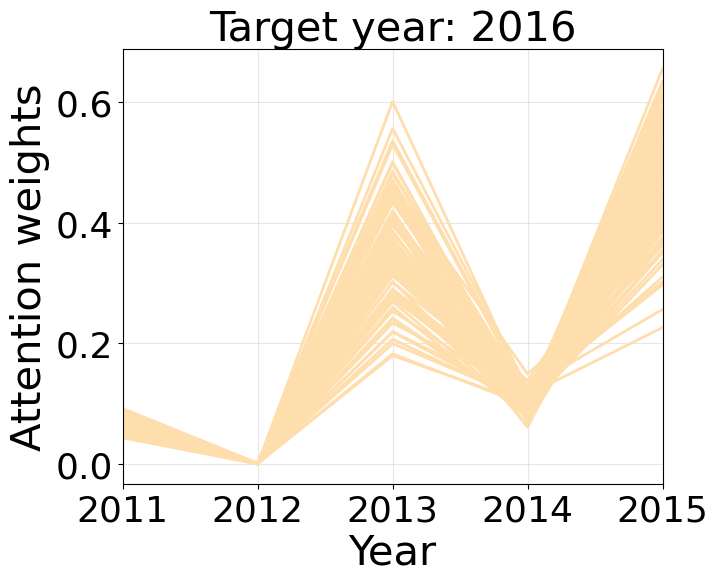}
}
\caption{Attention weights over historical years in the look-back window for different target years. The curves show the attention weights for  20\% randomly sampled counties.}
\label{fig:att}
\end{figure}

In Figure~\ref{fig:att}, we show the attention weights assigned to past years for each target year from 2013 to 2016. In most cases, the model places higher attention weights on more recent years, as they are more likely to share similar environmental and management conditions with the target year. In contrast, years such as 2011 and 2012 consistently receive low attention across target years (even when we set the next year 2013 as the target year). This is largely because these years (2011 and 2012) experienced frequent extreme weather events, making their temporal patterns less representative and therefore less useful for prediction in a different year.

\section{Conclusion}
In this paper, we propose a new framework to address two fundamental challenges in large-scale crop yield prediction:  (1) capturing both short-term and long-term temporal dynamics of crop growth, and (2) adapting to heterogeneous spatial regions. Our framework integrates a multi-scale temporal network LYRA with a retrieval-based strategy RaTAR to enhance the adaptability. 
Our evaluation on real-world county-level corn yield data demonstrates the effectiveness of both LYRA and RaTAR in improving crop yield prediction. We further show that the cross-year attention mechanism successfully identifies the most relevant historical years for each prediction. In addition, our analysis of the retrieval process reveals the benefits of the residual-based retrieval strategy and highlights the importance of the refinement step  in RaTAR.

Although the proposed method has been developed in the context of crop modeling, it can be generally applied to many other real-world earth science problems (e.g., prediction of greenhouse gas emissions, water quality monitoring) that exhibit complex temporal dynamics and high spatial heterogeneity. Our future work will further explore
the incorporation of domain knowledge (e.g., the mass balance of carbon cycles in each agricultural region) to further improve the retrieval and augmentation process.

\bibliographystyle{ACM-Reference-Format}
\bibliography{sample-base}

\appendix

\section{Dataset Details}
\label{appendix:dataset}

\begin{figure}[t]
  \centering
  \includegraphics[width=0.8\linewidth]{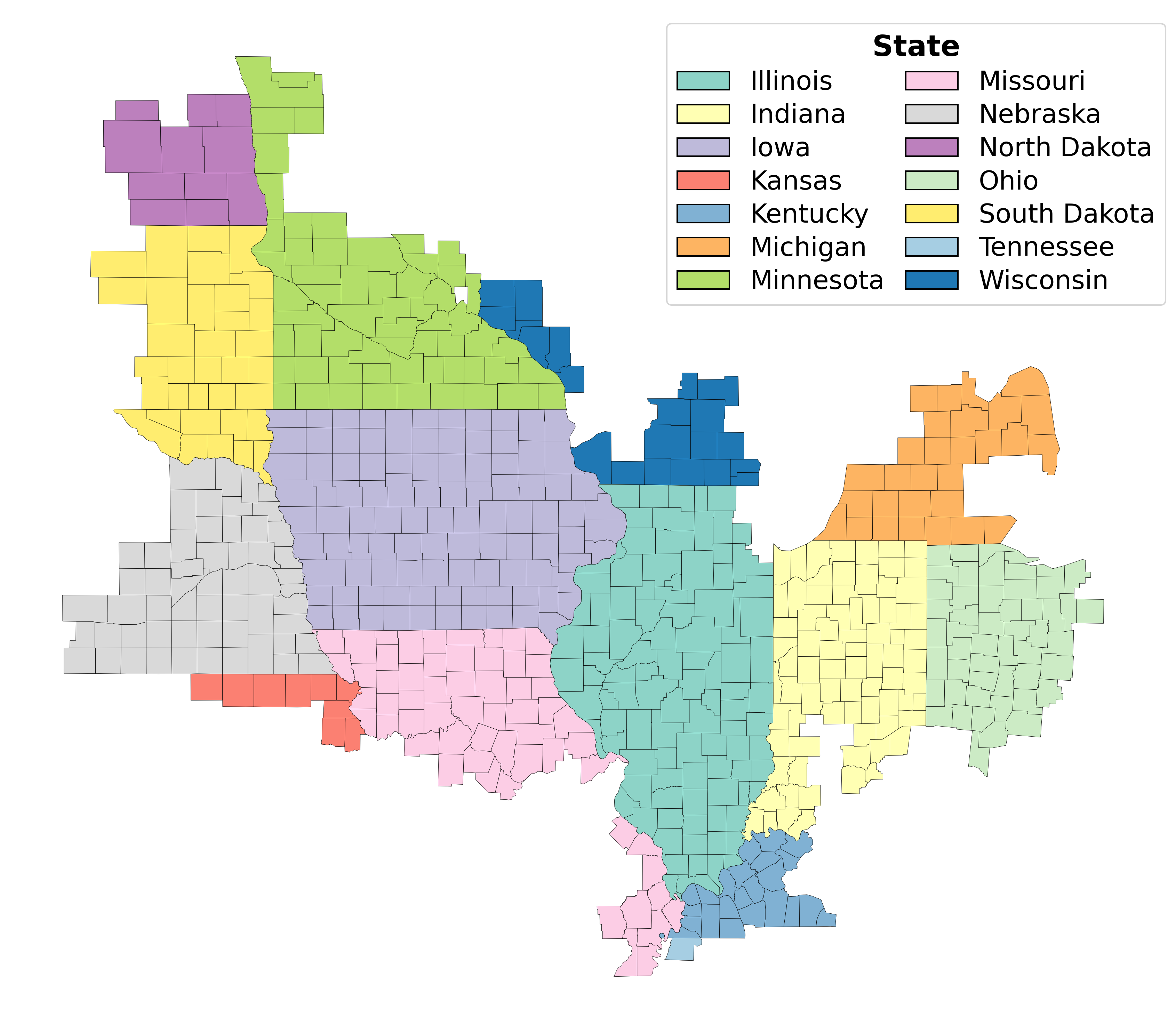}
  \Description{A map showing 630 counties across 14 U.S. states, highlighting the study region.}
  \vspace{-0.25in}
  \caption{Study region of 630 counties across 14 U.S. states.}
  \label{fig:state_map}
\end{figure}

\begin{table}[h]
\centering
\small
\begin{tabular}{p{1.4cm} p{4.5cm} p{1.5cm}}
\hline
\textbf{Variable} & \textbf{Description} & \textbf{Unit} \\
\hline
\multicolumn{3}{l}{\textit{Weather forcing}} \\
RADN & Downward solar radiation & MJ m$^{-2}$ d$^{-1}$ \\
TMAX\_AIR & Daily maximum temperature & $^\circ$C \\
TDIF\_AIR & Diurnal temperature range & $^\circ$C \\
HMAX\_AIR & Daily maximum humidity & kPa \\
HDIF\_AIR & Diurnal humidity range & kPa \\
WIND & Wind speed & km d$^{-1}$ \\
PRECN & Precipitation & mm d$^{-1}$ \\
\hline
\multicolumn{3}{l}{\textit{Soil / plant properties (0--30 cm gSSURGO)}} \\
Crop\_Type & Crop type (1 = corn, 5 = soybean) & -- \\
TBKDS & Bulk density & Mg m$^{-3}$ \\
TSAND & Sand content & g kg$^{-1}$ \\
TSILT & Silt content & g kg$^{-1}$ \\
TFC & Water content at field capacity & m$^3$ m$^{-3}$ \\
TWP & Water content at wilting point & m$^3$ m$^{-3}$ \\
TKSat & Saturated hydraulic conductivity & mm h$^{-1}$ \\
TSOC & Soil organic carbon & gC kg$^{-1}$ \\
TPH & Soil pH & -- \\
TCEC & Cation exchange capacity & cmol$^+$ kg$^{-1}$ \\
\hline
\multicolumn{3}{l}{\textit{Other variables}} \\
GPP & Daily gross primary productivity & gC m$^{-2}$ d$^{-1}$ \\
Year & Calendar year indicator & -- \\
\hline
\end{tabular}
\caption{Descriptions and units of the 19 input features used for county-level corn yield prediction.}
\label{tab:feature_definitions}
\end{table}

Our study region spans 630 counties across 14 states in the U.S. Corn Belt and surrounding areas (Figure~\ref{fig:state_map}). The map shows the county-level spatial distribution and state boundaries, highlighting the geographical breadth and environmental diversity of the dataset. The counties encompass a wide range of soil, climate, and management conditions, reflecting the substantial spatial heterogeneity that yield prediction models must handle. The dataset covers annual corn yield and corresponding environmental features from 2000 to 2020. Table~\ref{tab:feature_definitions} summarizes the 19 input features used in our experiments, including weather forcing variables, soil properties, vegetation productivity, and other auxiliary information.









\end{document}